\documentclass[sigconf]{acmart}
\renewcommand\footnotetextcopyrightpermission[1]{}
\usepackage{hyperref}
\hypersetup{colorlinks=true, linkcolor=blue, anchorcolor=blue, citecolor=blue}
\usepackage{cleveref}
\usepackage{array}
\usepackage{algorithm}
\usepackage{algorithmic}
\usepackage{multirow}
\usepackage{balance}

\AtBeginDocument{%
  \providecommand\BibTeX{{%
    \normalfont B\kern-0.5em{\scshape i\kern-0.25em b}\kern-0.8em\TeX}}}
\begin{document}
\fancyhf{} % 清空当前页眉和页脚
\pagestyle{empty} % 设置页面样式为空，禁用页眉和页脚

\title{StableMoFusion: Towards Robust and Efficient Diffusion-based Motion Generation Framework
}

% \author[1]{Yiheng Huang}
% \author[2]{Hui Yang}
% \author[3]{Chuanchen Luo}
% \author[2]{Yuxi Wang}
% \author[1]{Shibiao Xu}
% \author[4]{Zhaoxiang Zhang}
% \author[1]{Man Zhang\thanks{Corresponding authors: ManZhang (\texttt{zhangman@bupt.edu.cn}) and
% Junran Peng(\texttt{jrpeng4ever@126.com})
% }}
% \author[5]{Junran Peng
% % \textsuperscript{\Letter}}
% \textsuperscript{$\ast$}}
% \affil[1]{Beijing University of Posts and Telecommunications}
% \affil[2]{CAIR, HKISI, CAS}
% \affil[3]{Shandong University}
% \affil[4]{Institute of Automation, Chinese Academy of Science}
% \affil[5]{University of Science and Technology Beijin}

\author{Yiheng Huang}
\affiliation{
  \institution{Beijing University of Posts and Telecommunications}\country{}}
\email{hyh654@bupt.edu.cn}
\orcid{0009-0003-2972-8492}

\author{Hui Yang}
\affiliation{%
  \institution{CAIR, HKISI, Chinese Academy of Sciences$^{1}$}\country{}}
  % \streetaddress{1 Th{\o}rv{\"a}ld Circle}

\author{Chuanchen Luo}
\affiliation{%
  \institution{Institute of Automation, Chinese Academy of Sciences}\country{}\country{}}

\author{Yuxi Wang}
\affiliation{%
  \institution{CAIR, HKISI, Chinese Academy of Sciences$^{1}$}
  \thanks{1 Centre for Artificial Intelligence and Robotics, Hong Kong Institute of Science \& Innovation}\country{}}
  % \streetaddress{1 Th{\o}rv{\"a}ld Circle}

\author{Shibiao Xu}
\affiliation{%
  \institution{Beijing University of Posts and Telecommunications}\country{}}

\author{Zhaoxiang Zhang}
\affiliation{%
  \institution{Institute of Automation, Chinese Academy of Sciences}\country{}}

\author{Man Zhang}
\authornote{Corresponding author}
\affiliation{%
 \institution{Beijing University of Posts and Telecommunications}\country{}}
\email{zhangman@bupt.edu.cn}

\author{Junran Peng}
\authornote{Project Leader}
\affiliation{%
  \institution{University of Science and Technology Beijing}\country{}}
\email{jrpeng4ever@126.com}

\begin{abstract}

Thanks to the powerful generative capacity of diffusion models, recent years have witnessed rapid progress in human motion generation. Existing diffusion-based methods employ disparate network architectures and training strategies. The effect of the design of each component is still unclear. In addition, the iterative denoising process consumes considerable computational overhead, which is prohibitive for real-time scenarios such as virtual characters and humanoid robots. For this reason, we first conduct a comprehensive investigation into network architectures, training strategies, and inference process. Based on the profound analysis, we tailor each component for efficient high-quality human motion generation. Despite the promising performance, the tailored model still suffers from foot skating which is an ubiquitous issue in diffusion-based solutions. To eliminate footskate, we identify foot-ground contact and correct foot motions along the denoising process. By organically combining these well-designed components together, we present StableMoFusion, a robust and efficient framework for human motion generation. Extensive experimental results show that our StableMoFusion performs favorably against current state-of-the-art methods. Project page: \href{https://h-y1heng.github.io/StableMoFusion-page/}{https://h-y1heng.github.io/StableMoFusion-page/}.

\end{abstract}

\maketitle

\section{Introduction}

% 背景阐述
Human motion generation aims to generate natural, realistic, and diverse human motions, which could be used for animating virtual characters or manipulating embodied robots to imitate vivid and rich human movements without long-time manual motion modeling and professional skills\cite{azadi2023make,motiondiffuse,dabral2023mofusion}. It shows great potential in the fields of animation, video games, film production, human-robot interaction, and \textit{etc}. Recently, the application of diffusion models to human motion generation has led to significant improvements in the quality of generated motions~\cite{MDM,motiondiffuse,mld}.

% 动机：
Despite the notable progress made by diffusion-based motion generation methods, its development is still hindered by several fragmented and underexplored issues: 
1) \textbf{Lack of Systematic Analysis}: these diffusion-based motion generation work usually employ different network architectures and training pipelines, which hinders cross-method integration and the adoption of advancements from related domains.
%  makes it difficult to compare and analyze the design of each component in the diffusion pipeline between different methods.
2) \textbf{Long Inference Time}: due to the time-consuming iterative sampling process, most existing methods are impractical for applications with virtual characters and humanoid robots, where real-time responsiveness is crucial. 
3) \textbf{Footskate Issue}: foot skating in generated motions remains a major concern. This significantly undermines the quality of generated motions and limits their practical applicability.
% However, these diffusion-based motion generation work usually employ different network architectures and training pipelines, which makes it difficult to compare and analyze the design of each component in the diffusion pipeline between different methods. Furthermore, the application problems such as long inference time and foot slip phenomenon have not been fully solved.

\begin{table}[t!]
  \caption{StableMoFusion achieves superior performance. Lower FID and higher R Precision mean, the better.}
  \label{tab1: Comparison_with_SOTA}
  \centering
  \vspace{-0.2cm}
   \resizebox{\linewidth}{!}{
  \scriptsize
  
  \begin{tabular}{l | c | m{1.7cm}<{\centering}}
    \toprule
    Method &   FID$\downarrow$& R Precision (top3)$\uparrow$  \\
    \midrule
    MDM ~\cite{MDM} &  0.544 & 0.611  \\
    MLD ~\cite{mld} & 0.473 & 0.772\\
    MotionDiffuse~\cite{motiondiffuse} &  0.630  & 0.782 \\
    % T2M-GPT~\cite{zhang2023t2mgpt} &  0.116  & 0.775 \\
      ReMoDiffuse~\cite{zhang2023remodiffuse} &  0.103 & 0.795  \\
      StableMoFusion (Ours) &  \textcolor{red}{0.098} & \textcolor{red}{0.841} \\
    
    \bottomrule
  \end{tabular}
  }
\end{table}

\begin{figure}[t!]
  \centering
   \includegraphics[width=\linewidth]{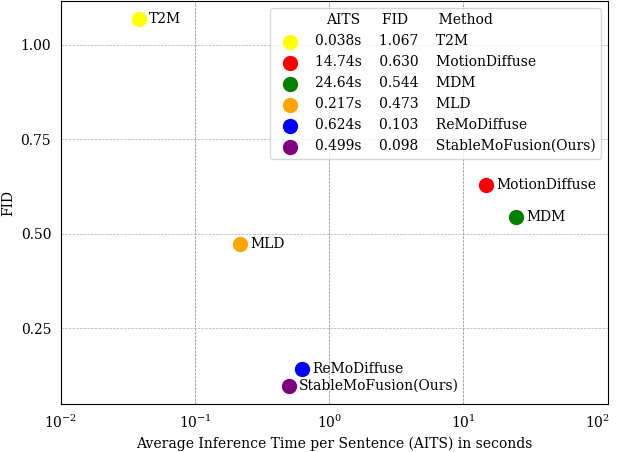}
   \caption{Comparison of the inference time costs on motion generation. The closer the model is to the origin, the better.}
   \label{fig:AITS}
\end{figure}

Therefore, in order to fill these research gaps and enhance the effectiveness and reliability of diffusion-based motion generation in practical applications, our study conducts a comprehensive and systematic investigation through network architectures, training strategies, and inference process, to reveal the influence of each factor. Ultimately, based on our
observations, we present a robust and efficient framework for diffusion-based motion generation, called \emph{\textbf{StableMoFusion}}, as illustrated in ~\autoref{fig:framework}.

In StableMoFusion, we use Conv1D UNet with AdaGN and linear cross-attention as the motion-denoising network and improve its generalization capability with GroupNorm tweak. During training, two effective strategies were employed to enhance the network's ability to generate motion. During inference, we use four training-free acceleration tricks to achieve efficient inference. Furthermore, we present a footskate cleanup method based on a mechanical model and optimization.

% 结果速览
Extensive experiments demonstrate that StableMoFusion achieves an excellent trade-off between text-motion consistency and motion quality compared to other state-of-the-art methods, as shown in ~\autoref{tab1: Comparison_with_SOTA}. Meanwhile, StableMoFusion's efficient inference process notably reduces the number of iterations required for generation from 1000 to 10, as well as shorter inference times than methods of about the same performance, achieving an average inference time of 0.5 seconds on the Humanm3D test set, as shown in ~\autoref{fig:AITS}. In addition, our footskate cleanup method within diffusion framework sizably solves the foot skating problem of motion generation as shown in ~\Cref{sec:Qualitative result}.

Our major contributions can be summarized as follows:
\begin{itemize}
\item We perform a systematic evaluation and analysis on the design of each component in the diffusion-based motion generation pipeline, including network architectures, training strategies, and inference process. 
% To our best knowledge, this is the first work for surveying diffusion framework for motion generation, providing valuable practical guidelines for future research.

\item We propose an effective mechanism to eliminate foot skating which is a comment issue in current methods. 
% The paper addresses two practical challenges: time-consuming inference and foot skating, which are crucial problems in industry. 
% By offering solutions to these challenges, the work enables the practical deployment of diffusion-based motion generation methods across various multimedia contexts.

\item By consolidating these well-designed components, we present a robust and efficient diffusion-based motion generation framework named \emph{StableMoFusion}. Extensive experiments demonstrate its superiority in text-motion consistency and motion quality.
\end{itemize}

\section{Related Work}
\subsection{Motion Diffusion Generation}

Recent advancements in diffusion models have greatly enhanced human motion generation quality. Notable works include MotionDiffuse~\cite{motiondiffuse}, which integrates text features via cross-attention; MDM~\cite{MDM}, exploring various denoising networks with Transformer or GRU; PhysDiff~\cite{yuan2023physdiff}, adding realism with physical constraints; MLD~\cite{mld}, accelerating generation using VAE latent spaces; and ReMoDiffuse~\cite{zhang2023remodiffuse}, enhancing with retrieval mechanisms. MotionLCM~\cite{dai2025motionlcm} achieves real-time controllable motion generation through latent consistency model.

However, the field of motion generation is still grappling with a lack of consistent and fair comparisons due to varying network architectures, sampling methods, training strategies, etc. Our work addresses this by providing a thorough exploration of these aspects, revealing the influence of each factor.

% \subsection{Training-Free Sampling}
% To reduce the inference time with a trained network, there have been many advanced samplers to accelerate DDPM~\cite{ddpm}.

% Song et al.~\cite{SDE} show that using Stochastic Differential Equation (SDE) for sampling has a marginally equivalent probability Ordinary Differential Equations (ODE). And then, DDIM~\cite{ddim} constructs a class of non-Markovian diffusion processes that realize skip-step sampling. PNDN~\cite{liu2021pndm} uses pseudo numerical to accelerate the deterministic sampling process. DEIS~\cite{zhang2022deis} and DPMSolver~\cite{lu2022dpm} improve upon DDIM by numerically approximating the score functions within each discretized time interval. 

% Meanwhile, several work have focused on speeding up stochastic sampling. For example, Gotta Go Fast~\cite{jolicoeur2021gotta} utilizes adaptive step sizes to speed up SDE sampling, and Lu et al.~\cite{lu2022dpm++} converts the higher-order ODE solver into an SDE sampler to address the instability issue.

% While these samplers have demonstrated efficacy in image generation, their impact on motion diffusion models remains unexplored. In this work, we evaluate them to find the most appropriate one for motion generation.

\subsection{Footskate Cleanup}
In order to generate realistic motions in computer animation, various methods have been developed to improve footskate issue.
% Footskate cleanup refers to a technique used in computer graphics and animation to address the issue of unnatural foot movements, where the feet of characters in animations appear to slide unnaturally across surfaces instead of making realistic contact and interaction. This phenomenon can significantly degrade the quality and realism of animations, especially in scenarios where characters are walking, running, or interacting with the environment. 

Edge~\cite{tseng2022edge} embeds the foot contact term into the action representation for training and applies Contact Consistency Loss as a constraint to keep the physical plausibility of motion. RFC~\cite{yuan2020residual}, Drop~\cite{jiang2023drop} and Physdiff~\cite{yuan2023physdiff} use reinforcement learning to constrain the physical states of actions, such as ground force reaction and collision situations to get a realism motion. UnderPressure~\cite{Mourot22} and GroundLink~\cite{han2023groundlink} respectively collect foot force datasets during motion. UnderPressure~\cite{Mourot22} also utilizes this dataset to train a network capable of predicting vertical ground reaction forces. 

Based on this, Our approach uses a novel module to optimize the sliding of the weighted feet in diffusion-generated motions during inference by backpropagation. Compared to traditional feet-joint loss during training, our offline-extendable and force-based method is more stable and reliable for the animation industry.

% framework figure of StabelMoFusion
\begin{figure*}[t!]
  \centering
 \includegraphics[width=\linewidth]{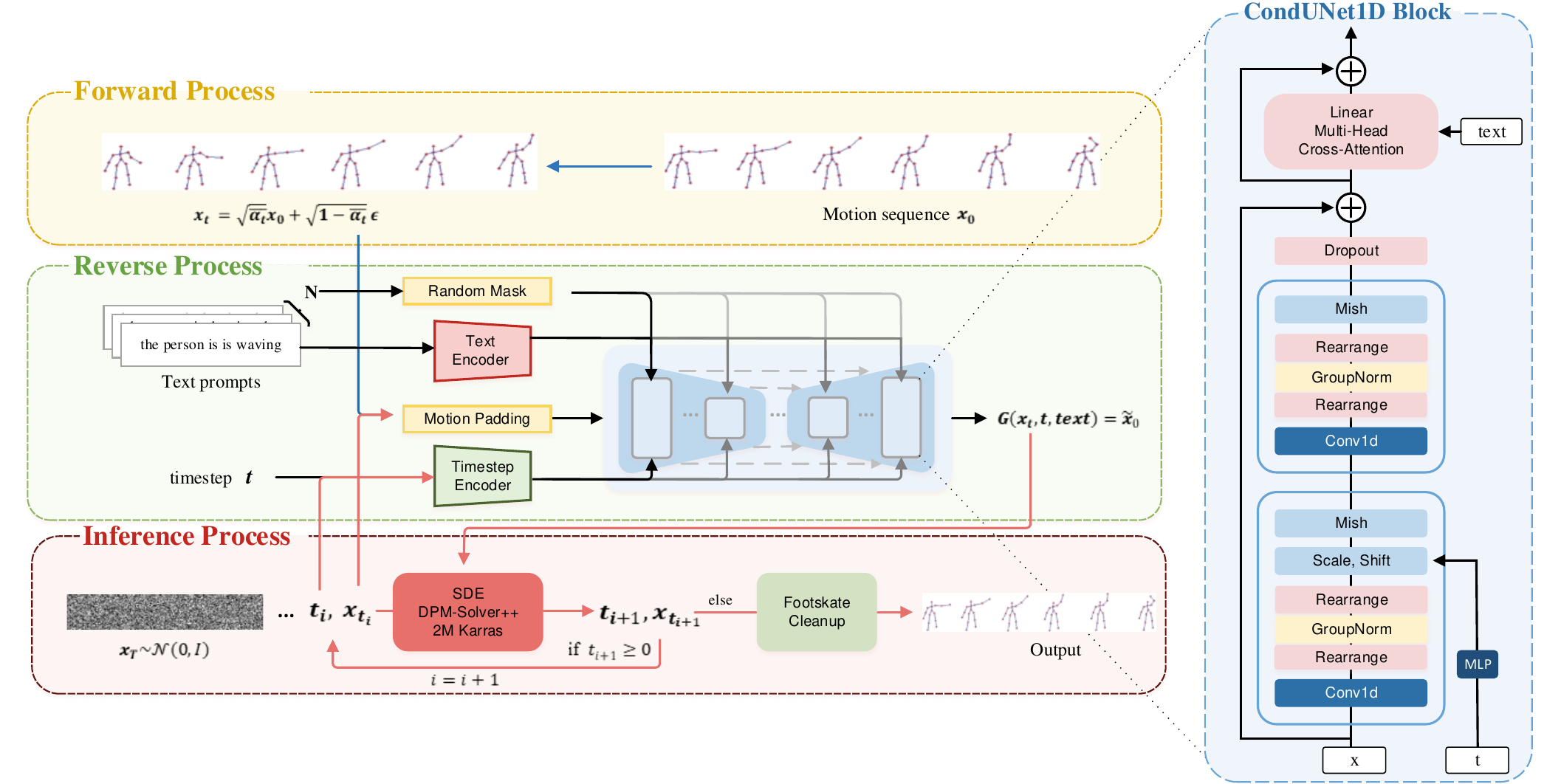}
   \caption{Overview of StableMoFusion, which is composed of a diffusion forward process, a reverse process on CondUNet1D motion-denoising network, and an efficient inference. The colors of the arrows indicate different stages: blue for training, red for inference, and black for both. }
   \label{fig:framework}
\end{figure*}

\section{Preliminaries}\label{sec: Preliminaries}

The pipeline of Diffusion model~\cite{ddpm} involves three processes: a \textbf{forward process} that gradually diffuses noise into the sample, a \textbf{reverse process} that optimizes a network to eliminate the above perturbation from noisy samples, and an \textbf{inference process} that utilizes the trained network to iteratively denoise noisy samples.

Specifically, $x_0 \in R^{N \times M}$ represents a sequence of human poses for N frames, where M is the dimension of human pose representations. Then randomly select a timestep $t\sim U [0, T]$ and the noisy motion $x_t$ after t-step diffusion is gained by ~\autoref{eq:diffusion forward},
\begin{equation}\label{eq:diffusion forward}
    x_t =  \sqrt{\bar{\alpha_t}} x_0 + \sqrt{1-\bar{\alpha_t}} \epsilon
\end{equation}
where $\epsilon$ is a Gaussian noise. $\sqrt{\bar{\alpha_t}}$ and $\sqrt{1-\bar{\alpha_t}}$ are the strengths of signal and noise, respectively. $\alpha_t$ decreases as t increases.
When $\sqrt{\bar{\alpha_t}}$ is small enough, we can approximate $x_t \sim \mathcal N (0, {I})$.

Next, given a motion-denoising model $\mathbf{G_{\theta}}(x_t, t, c)$ for predicting the original sample, parameterized by $\theta$, the optimization can be formulated as follows: 
\begin{equation}\label{eq: simple objective1}
    \mathop{\min}_{\theta} E_{t\sim U[0,T],x_0\sim p_{data}}
    ||\mathbf{G}_{\theta}(x_t, t, c) - x_0||_2^2
\end{equation}

% In summary, the template for training the denoising network using both forward process and reverse process is shown in ~\Cref{alg1}.

% \begin{algorithm}[h]
%     \caption{Training}
%     \label{alg1}
%     \begin{algorithmic}
%         \STATE Initialize $i=0$
%         \REPEAT 
%         \STATE $\mathbf{x}_0 \sim q(x_0|c)$
%         \STATE $t \sim Uniform{\left\{0,\cdots,T \right\}}$
%         \STATE $\mathbf{\epsilon} \sim \mathcal N (0, \mathbf{I})$
%         \STATE Take AdamW gradient descent  step on
%         \STATE $\hspace{\algorithmicindent}\bigtriangledown_{\theta}||\mathbf{G}_{\theta}(\sqrt{\bar{\alpha_t}} \mathbf{x}_0 + \sqrt{1-\bar{\alpha_t}} \mathbf{\epsilon}, t, c) - \mathbf{x}_0||_2^2$
%         \STATE $i \leftarrow i + 1$
%         \UNTIL{$i=50,000$}
%     \end{algorithmic}
% \end{algorithm}

In the inference process, a trained motion-denoising network can progressively generate samples from noise with various samplers. For instance, DDPM \cite{ddpm} iteratively denoise the noisy data from $t$ to a previous timestep $t^{\prime}$, as shown in ~\Cref{alg2}.  
% $ x_{t-1}$ will be fed into $\mathbf{G_{\theta}}(x_t, t, c)$ again until $t-1$ becomes 0. 
% % https://zhuanlan.zhihu.com/p/655568910
% % formula (7) from https://arxiv.org/pdf/2006.11239.pdf
% Most methods (\textit{\textit{e.g.}} MDM~\cite{MDM}, MotionDiffuse~\cite{motiondiffuse}) use DDPM~\cite{ddpm} sampler for inference, and its inference process is . 
% However, the 1000-step iteration leads to time-consuming inference. For this reason, recently, there have been many advanced samplers to reduce inference time by minimizing the denoising step. While such samplers are common in image generation, their application in motion generation has been relatively rare.

\begin{algorithm}[h]
    \caption{Inference}
    \label{alg2}
    \begin{algorithmic}
        \STATE Given a text prompt $c$
        \STATE $\mathbf{x}_{t} \sim \mathcal N (0, \mathbf{I})$
        \FOR{$t=T$ to $1$}
            \STATE  $ \widetilde{\mathbf{x}}_0 = \mathbf{G}(\mathbf{x}_{t}, t, c)$
            % \STATE Sample $\mathbf{x}_{t-1} \sim q\left(\mathbf{x}_{t-1} \mid \mathbf{x}_t, \widetilde{\mathbf{x}}_0\right)$ :
            \STATE $\epsilon \sim \mathcal N (0, I)$ if $t>1$, else $\epsilon=0$
            \STATE $\mathbf{x}_{t-1} = \frac{\sqrt{\bar{\alpha}_{t-1}} \beta_t}{1-\bar{\alpha}_t}\widetilde{\mathbf{x}}_0+\frac{\sqrt{\alpha_t}\left(1-\bar{\alpha}_{t-1}\right)}{1-\bar{\alpha}_t} \mathbf{x}_t + \tilde{\beta}_t \epsilon $
        \ENDFOR
        \RETURN $\mathbf{x}_0$
    \end{algorithmic}
\end{algorithm}

% block visualization figure
\begin{figure*}[t!]
  \centering
  \includegraphics[width=\linewidth]{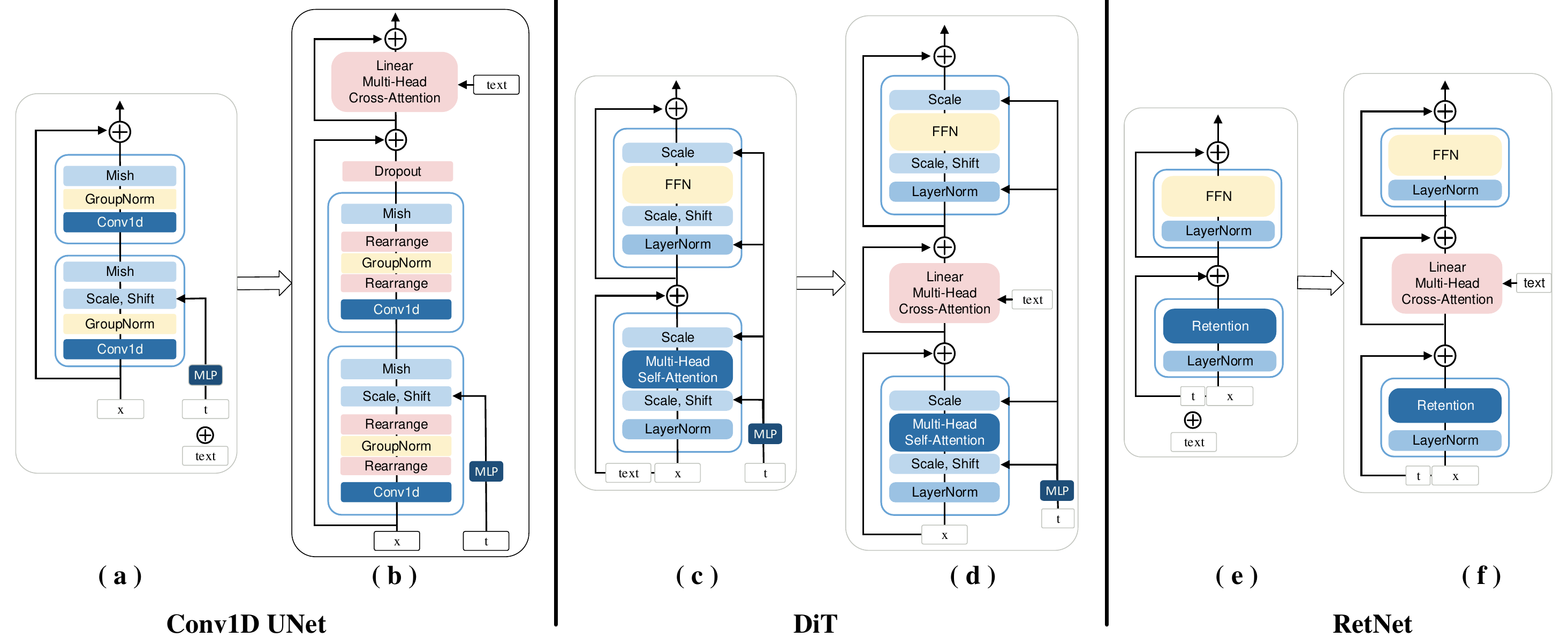}
   \caption{Visualization of the block structure and their adjustments of Conv1D UNet, DiT and RetNet. Pink blocks indicate structures that have been added or modified. }
   \label{fig:block visualization}
\end{figure*}

\section{Method}
Through comprehensive exploratory experiments conducted on diffusion-based motion generation, we propose a novel diffusion framework, named StableMoFusion, as illustrated in \autoref{fig:framework}, to facilitate robust and efficient motion generation. This section begins with our investigation of the architecture of motion-denoising networks. Next, we discuss several training strategies pivotal for enhancing model performance in ~\Cref{sec:training strategies}. Subsequently, we introduce our improvements in the inference process in \Cref{sec:efficient inference}, tailored to enable efficient inference. Lastly, we discuss and present a solution to the footskate issue in \Cref{sec: footskate}.

\subsection{Model Architecture}\label{sec:Model Architecture}

In this subsection, we explore three motion denoising network architectures for text-to-motion generation tasks: Conv1D UNet~\cite{dabral2023mofusion,gmd}, Diffusion Transformer (DiT)~\cite{dit} and the latest Retentive Network (RetNet)~\cite{sun2023retentive}.

\subsubsection{Conv1D UNet}

\paragraph{Baseline} We choose the Conv1D UNet with AdaGN~\cite{adagn} as the baseline, where each block is shown as \autoref{fig:block visualization} (a). The skip connections and layer-by-layer reconstruction mechanism of UNet, as shown in \autoref{fig:framework}, are well-suited for retaining spatial and temporal details in motion sequences. Moreover, UNet can extract and reconstruct features at different scales, producing more realistic motion sequences. 

In the Conv1d block, the Clip-based~\cite{radford2021clip} sentence-level textural embedding, $c_{text} \in R^{D}$, $D$ is the dimension of text, 
is projected into $scale$ and $shift$ along with timestep embedding 
 for feature-wise linear modulation (FiLM): $x^{(i)} \times (1 + scale) + shift$, $x^{(i)}$ is the ith-frame pose embedding. That means a consistent condition-based linear mapping is applied to the pose embedding of each frame. 
This method is well-suited for injecting timestep into noisy motion for denoising, but falls short This is because the entire motion sequence is subjected to the same timestep-level noise perturbation, while each frame of the motion sequence involves distinct semantics within the text sentence.

% and modify the structure to a canonical UNet structure, which consists of four downsampling stages. The motion length $n$ is successively reduced from $N$to $\left \lfloor N/8 \right \rfloor$, and then the corresponding up-sampling phase is used to up-sample. There are two residual Conv1D blocks for each down-sampling or up-sampling stage, with a single block shown as \autoref{fig:block visualization} (a).

% \vspace{-0.3cm}
\paragraph{Block Adjustment}\label{para: block adjustment}  
% We introduce Residual Cross-Attention into the Conv1D block to attentively adjust motion embedding (as Query) based on word-level semantic information (as Key and Value), as shown in ~\autoref{fig:block visualization} (b). 
% In addition, dropout is incorporated into the original Conv1D block to enhance model generalization.
% Compared to sentence-level embedding in the baseline, we employ additional transformer encoder layers that follow CLIP to convert word-level text embeddings into motion modality-related representations, $text \in R^{L \times D}$, $L$ is the length of text, for cross-modality attention.

We integrated Residual Cross-Attention into the Conv1D block for alignment of motion embeddings (as Query) with word-level semantics (as Key and Value). In addition, dropout is added to boost model generalization, as depicted in \autoref{fig:block visualization} (b). 
Beyond the baseline's sentence-level embedding, we employ additional transformer layers that follow CLIP to refine word-level text embeddings into motion-related representations, $c_{text} \in R^{L \times D}$, $L$ is the length of text, for effective cross-modality attention.

\paragraph{GroupNorm Tweak} 
We rearranged the dimensions of motion embedding to enable feature-wise Group Normalization, as depicted in \autoref{fig:block visualization} (b).
Given that Conv1D performs convolutions along the sequence dimension, directly applying group normalization to its output is akin to sequence-wise normalization, which is highly sensitive to the proportion of sequence padding. Consequently, the baseline model experiences a significant performance drop on datasets with diverse motion lengths, such as KIT-ML, due to the increased impact of padding.

\subsubsection{Diffusion Transformer} 

% The primary feature of DiT compared to transformer encoders lies in its incorporation of a scale and shift adjustment using timesteps before and after each autoregressive computation. This ensures that the denoising process remains closely aligned with the timesteps, thereby mitigating excessive deviations in the motion denoising trajectory during the autoregressive computation. Therefore, in this work, we apply the foundational architecture principles of the Diffusion Transformer to motion generation tasks, to explore its efficacy and potential enhancements for generating natural motion sequences.

\paragraph{Baseline} To explore the effectiveness of the DiT structure for motion generation, we replace the Vision Transformer used for images in the DiT with self-attention used for motion data as the baseline, with the basic block structure shown in \autoref{fig:block visualization} (c). For text-to-motion generation, we embed text prompts via the CLIP~\cite{radford2021clip} encoder and project them into tokens concatenated with motion embeddings for self-attention. It scales and shifts the motion embedding before and after each autoregressive computation using timestep, which ensures the motion-denoising trajectory is closely aligned with the timestep.

\paragraph{Block Adjustment} We have also tried to incorporate Linear Multi-Head Cross-Attention into the DiT framework, as shown in ~\autoref{fig:block visualization} (d). 
% Fusing all the text information into the one-dimensional text embedding may lead to insufficient bootstrapping information, 
This adjustment allows for a more nuanced fusion of textual cues with motion dynamics than fusing all the text information into the one-dimensional text embedding in baseline, which enhances the coherence and relevance of generated motion sequences.

\subsubsection{Retentive Network} 

% In our study, we explore the application of the Retentive Network (RetNet) in the motion generation, a recently introduced autoregressive framework tailored for time series data. RetNet offers several advantages over conventional Transformer architectures, including concurrent training, reduced computational overhead, and notable performance gains. Notably, RetNet incorporates a gated multi-scale retention mechanism, which enhances information retention and processing capabilities, thereby enabling nuanced comprehension and generation of motion sequences. Through our investigation, we aim to ascertain the feasibility of leveraging RetNet for motion generation tasks.

\paragraph{Baseline} Our RetNet baseline follows a straightforward implementation similar to MDM, where the timesteps encoding is concatenated with the textual projection to form tokens, which are then fed along with motion embeddings into RetNet, with its basic block shown in ~\autoref{fig:block visualization} (e). RetNet incorporates a gated multi-scale retention mechanism, which enhances information retention and processing capabilities, thereby enabling nuanced comprehension and generation of motion sequences. Through our investigation, we aim to ascertain the feasibility of leveraging RetNet for motion generation tasks.

\paragraph{Block Adjustment} 
To further integrate textual information, we also employ Linear Multi-Head Cross-Attention between retention and FFN, as shown in ~\autoref{fig:block visualization} (f). By segregating temporal and textual features, our approach aims to preserve the distinct characteristics of each modality and allow the model to independently learn and leverage relevant cues for motion generation. This separation enhances the model's interpretability and flexibility, enabling it to better capture the intricacies of both temporal dynamics and semantic context.

\subsubsection{Final Model Architecture} \quad

Ultimately, we choose the Conv1D UNet with block adjustment and GroupNorm tweak as the motion-denoising model of StableMoFusion, as shown in ~\autoref{fig:framework}. We call this network as CondUNet1D.
% The self-attention enables the global receptive field along the temporal dimension, which benefits the modeling of long-range dependency.
Both DiT and RetNet use the idea of attention to activate the global receptive field in the temporal dimension, which benefits the modeling of long-range dependency. The receptive field of Conv1D UNet is mainly in the convolution kernel window, promoting a coherent and smooth transition between frames. We tend to prioritize smoother generation in current applications of motion generation.

In our StableMoFusion, we set the base channel and channel multipliers of UNet to 512 and [2,2,2,2] respectively. For text encoder, we leverage pre-trained CLIP \cite{radford2021clip} token embeddings, augmenting them with four additional transformer encoder layers, the same as MotionDiffuse \cite{motiondiffuse}, with a latent text dimension of 256. For timesteps encoder, it is implemented using position encoding and two linear layers, the same as StableDiffusion~\cite{rombach2022high}, with a latent time dimension of 512. 

\subsection{Training Strategies}\label{sec:training strategies} 
Recent research has shown that training strategies in diffusion models are also crucial for generation capabilities \cite{cao2022survey}. In this subsection, we will analyze the impact of two empirically valid training strategies on diffusion-based motion generation: exponential moving average and classifier-free guidance.

\subsubsection{Exponential Moving Average} \quad

Exponential Moving Average (EMA) calculates a weighted average of a series of model weights, giving more weight to recent data. Specifically, assume the weight of the model at time t as $\theta_t$, then the EMA formulated as: $v_t=\beta \cdot v_{t-1}+(1-\beta) \cdot \theta_t$, where $v_t$ denotes the average of the network parameters for the first t iterations ($v_0 = 0$), and $\beta$ is the weighted weight value. 

During the training of the motion-denoising network, the network parameters change with each iteration, and the motion modeling oscillates between text-motion consistency and motion quality. Therefore, the use of EMA can smooth out the change process of these parameters, reduce mutations and oscillations, and help to improve the stability ability of the motion-denoising model.

\subsubsection{Classifier-Free Guidance} \quad

To further improve the generation quality, we use Classifier-Free Guidance (CFG) to train the motion-denoising generative model.
By training the model to learn both conditioned and unconditioned distributions (e.g., setting c = $ \emptyset $ for 10\% of the samples), CFG ensures that the models can effectively capture the underlying data distribution across various conditions. In inference, we can trade-off text-motion consistency and fidelity using s:
\begin{equation}\label{eq:cfg}
     G_s\left(x_t, t, c\right)=G\left(x_t, t, \emptyset\right)+s \cdot\left(G\left(x_t, t, c\right)-G\left(x_t, t, \emptyset\right)\right)
\end{equation}
This ability to balance text-motion consistency and fidelity is crucial for producing varied yet realistic outputs, enhancing the overall quality of generated motion.

\subsection{Efficient Inference}\label{sec:efficient inference}

Time-consuming inference time remains a major challenge for diffusion-based approaches. To address this problem, we improve inference speed by integrating four efficient and training-free tricks in the inference process: 1) efficient sampler, 2) embedded-text cache, 3) parallel CFG computation, and 4) low-precision inference.
% We chose MotionDiffuse ~\cite{motiondiffuse} as our starting point since its framework is cleaner than other methods without additional geometric constraints and complex design and it is more suitable for analyzing our various improvements. 

% \paragraph{Sampling Optimization}\label{sec: Sampler optimization}
% Sampling optimization aims to utilize a more advanced sampler on a pre-trained model for fewer steps. 

\subsubsection{Efficient Sampler} \quad

We replace DDPM with a more efficient sampling algorithm, DPM-Solver++~\cite{lu2022dpm++}, to reduce denoising iterations during inference. 
Specifically, we utilize the SDE variant of second-order DPM-Solver, a high-order solver for diffusion stochastic differential equations (SDEs). It introduces additional noise during the iterative sampling. Thereby, the stochasticity of its sampling trajectories helps to reduce the cumulative error~\cite{xu2023restart}, which is crucial for the realism of generated motion. 
In addition, we adopt the Karras Sigma~\cite{ODE} to set discrete timesteps. This method leverages the theory of constant-velocity thermal diffusion to determine optimal timesteps, thereby maximizing the efficiency of motion denoising within a given number of iterations.

\subsubsection{Embedded-text Cache} \quad

We integrate the Embedded-text Cache mechanism into the inference process to avoid redundant calculations. In diffusion-based motion generation, the text prompts remain unchanged across iterations, resulting in the same embedded text in each computation of the denoising network.
Specifically, we compute the text embedding initially and subsequently utilize the embedded text directly in each network forward, thereby reducing computational redundancy and speeding up inference.

\subsubsection{Parallel CFG Computation} \quad

We implement the inference process of CFG in parallel to speed up the single iteration calculation while maintaining model generation performance.
Due to the CFG mechanism ~\autoref{eq:cfg}, in each iterative step during inference, it is necessary to execute conditional and unconditional denoising, respectively, using the trained motion network, and then sum up the results.

\subsubsection{Low-precision Inference} \quad

We utilize half-precision floating point (FP16) computation during inference to accelerate processing. Newer hardware supports enhanced arithmetic logic units for lower-precision data types. By applying parameter quantization, we convert FP32 computations to lower-precision formats, effectively reducing computational demands, parameter size, and memory usage of the model.

\subsection{Footskate Reduction}\label{sec: footskate}

\begin{figure}[t!]
  \centering
   \includegraphics[width=\linewidth]{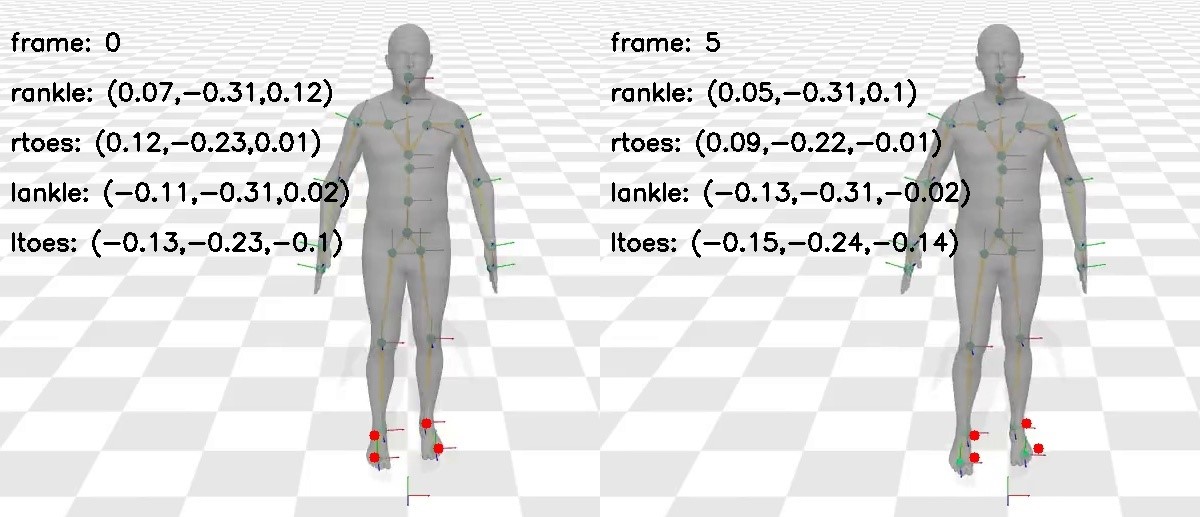}
   \caption{
   Footskate for a 20fps motion, with positional offsets of both feet from frame 0 (red) to frame 5 (green).}
   \label{fig:footskate_define}
\end{figure}
% \begin{figure}[ht!]
%   \centering
%    \includegraphics[width=\linewidth]{Fig/compare_3.png}
%    \caption{(a) Footskate motion. (b) Footskate cleanup motion by UnderPressure. Orange bounding boxes exhibit noticeable deformations. (c) Footskate cleanup motion by our method. The caption of motion is "kick left leg step back". Red points indicate the positions of right foot in the 8th frame. The image indicates the frame ID and the positions of 4 foot joints.}
%    \label{fig:footskate_compare_underpressure}
% \end{figure}

Foot skating is an animation artifact where a character's feet move unrealistically, often sliding across the ground, despite the character's body remaining in a consistent pose.
~\autoref{fig:footskate_define} illustrates an instance of the foot skating phenomenon. The depicted frames, separated by a 0.25-second interval, show the same posture but with both feet having shifted positions. Ideally, for this motion, we aim for the feet to maintain a consistent anchor point. Commonly, selecting the foot position from the midpoint of a foot skating sequence as fixed helps to minimize disruptions to surrounding frames.

The key to eliminating foot skating is to first identify the foot joints and frame ranges where foot skating occurs, and then anchor those keypoints at their positions $p$ in the intermediate frames. We formulate this constraint as a loss term shown in ~\autoref{eq: foot loss} where $j$ indicates joint and $f$ is frame ranges.  
\begin{equation}\label{eq: foot loss}
    L_{foot} =  \sum_j^{J_{skating}} \sum_f^{F_{skating}} (P_j - p)\\
\end{equation}

$J_{skating}$ contains all the joints where foot skating may occur, specifically including the right ankle, right toes, left ankle, and left toes. $F_{skating}$ is a collection of all frame ranges of the joint $j$. $P_j$ means the positions of joint $j$. We incorporate ~\autoref{eq: foot loss} to a gradient descent algorithm to correct foot skating motion.

Following UnderPressure~\cite{Mourot22}, we use vertical ground reaction forces (vGRFs) to identity foot joint $j$ and its skating frames $f$. The vGRFs prediction model of UnderPressure $V_{23}$ requires motion of a predefined 23 joints skeleton $S_{23}$, which is different from our motion data. In our work, we utilize HumanML3D\cite{t2m} with 22 skeletal joints $S_{22}$ and KIT-ML~\cite{plappert2016kit} motion with 21 skeletal joints. The subsequent foot skating cleanup primarily focused on HumanML3D. We transferred the pre-trained weights of $V_{23}$ to our own model $V_{22}^{\theta}$ using the constraints ~\autoref{eq: vGRFs transfer}, enabling us to directly predict the vertical ground reaction forces for HumanML3D motions. $P$ is keypoints of HumanML3D motion. $P_{S_{23}}$ is the result of retargeting $P$ to skeleton $S_{23}$. 

% \begin{equation}\label{eq: vGRFs transfer}
%   \mathop{\min}_{\theta} \|V_{22}^{\theta}(P) - V_{23}(P_{S_{23}}) \|_2^2   
% \end{equation}
\begin{gather}
 \mathop{\min}_{\theta} \|V_{22}^{\theta}(P) - V_{23}(P_{S_{23}}) \|_2^2 \label{eq: vGRFs transfer}\\
L = \omega_q L_{\text{pose}} + \omega_f L_{\text{foot}} + \omega_t L_{\text{trajectory}} + \omega_v L_{\text{vGRFs}} \label{eq: footskate loss}\\
  L_{\text{foot}} = L_{\text{foot}}(P, \hat{P}, V_{23}, P_{S_{23}})  \label{eq: footskate loss 1}    \\
   L_{\text{vGRFs}} = L_{\text{vGRFs}}(P, \hat{P}, V_{22}^{\theta}) \label{eq: footskate loss 2}   
\end{gather}

Drawing inspiration from UnderPressure~\cite{Mourot22}, we use foot contact loss $L_{foot}$ to fix contact joints, pose loss $L_{pose}$ and trajectory loss $L_{trajectory}$ to keep the semantic integrity of motion, vGRFs loss $L_{vGRFs}$ to keep valid foot pose. Our supplementary material provides detailed definitions of these loss terms. The final definition of our loss function is as ~\autoref{eq: footskate loss}~\cite{Mourot22} where $\omega_q$, $\omega_f$, $\omega_t$, $\omega_v$ are weights of its loss item. $P$ is keypoints of footskating motion and $\hat{P}$ is the result keypoints after footskate cleanup.

% \begin{equation}\label{eq: footskate loss}
%     L = \omega_q L_{\text{pose}} + \omega_f L_{\text{foot}} + \omega_t L_{\text{trajectory}} + \omega_v L_{\text{vGRFs}} 
% \end{equation}

% \begin{equation}\label{eq: footskate loss 1}
%   L_{\text{foot}} = L_{\text{foot}}(P, \hat{P}, V_{23}, P_{S_{23}})  
% \end{equation}
% \begin{equation}\label{eq: footskate loss 2}
%   L_{\text{vGRFs}} = L_{\text{vGRFs}}(P, \hat{P}, V_{22}^{\theta})   
% \end{equation}

Through our method, the footskate cleanup process can be generalized to various skeletal motions.

In a few cases, motion corrected by ~\autoref{eq: footskate loss} may occur unreasonable or unrealistic poses. The diffusion model trained on a large amount of motion data learns the prior knowledge of real motions and has the ability to correct the invalid motions. 

Therefore, we use our pretrained diffusion model to correct such cases. Motivated by OmniControl~\cite{xie2024omnicontrol} and Physdiff~\cite{yuan2023physdiff}, we incorporate footskate cleaning method into the diffusion denoising process, denoted as StableMoFusion$^*$.

% Quantitative results Comparison  whis SOTA
\begin{table*}[t!]
 \caption{Quantitative results on the HumanML3D test set. The right arrow  $\rightarrow$ means the closer to real motion the better. \textcolor{red}{Red} and \textcolor{blue}{Blue} indicate the best and the second best result. }
\label{tab4: benchmark HumanML3D}
  \centering
  
\resizebox{\linewidth}{!}{
  % \begin{tabular}{l c  m{1.8cm}<{\centering} c m{1.8cm}<{\centering}  m{1.5cm}<{\centering} cc}
  \begin{tabular}{lcccccc}
    \toprule
    \multirow{2}{*}{\centering Method} & 
    \multirow{2}{*}{\centering FID $\downarrow$} & 
    \multicolumn{3}{c}{R Precision$\uparrow$ }& 
    \multirow{2}{*}{\centering Diversity $\rightarrow$}  & 
    \multirow{2}{*}{\centering Multi-modality $\uparrow$} 
    \\
    % \multirow{2}{*}{\centering Inference Steps $\downarrow$} \\
    \cline{3-5}
    && top1 & top2 & top3 & & \\
    \midrule
     Real & $0.002^{ \pm .000}$ & $0.511^{ \pm .003}$ & $0.703^{ \pm .003}$ & $0.797^{ \pm .002}$  & $9.503^{ \pm .065}$ & - \\
    \midrule
    % JL2P ~\cite{ahuja2019language2pose}  & $11.02^{ \pm .046}$& $0.246^{ \pm .001}$ & $0.387^{ \pm .002}$ & $0.486^{ \pm .002}$ &  $7.676^{ \pm .058}$ & - & - \\
    % Text2Gesture ~\cite{bhattacharya2021text2gestures} & $7.664^{\pm{.030}}$ & $0.165^{\pm{.001}}$ & $0.267^{\pm{.002}}$ &$0.345^{\pm{.002}}$ & $6.409^{\pm{.071}}$ & - & -\\
    % TEMOS ~\cite{petrovich2022temos} & $3.734^{\pm{.028}}$ & $0.424^{ \pm .002}$ & $0.612^{ \pm .002}$ & $0.722^{\pm{.002}}$  &$8.973^{\pm{.071}}$ & $0.368^{\pm{.018}}$ & -  \\
     T2M ~\cite{t2m} & $1.067^{\pm{.002}}$ &  $0.457^{\pm{.002}}$&  $0.639^{\pm{.003}}$ & $0.743^{\pm{.003}}$ & $9.188^{\pm{.002}}$ & $2.090^{\pm{.083}}$  \\
    MDM ~\cite{MDM} & $0.544^{\pm{.044}}$ & $0.320^{ \pm .005}$ & $0.498^{ \pm .004}$ & $0.611^{\pm{.007}}$ & \textcolor{blue}{9.599}$^{\pm{.086}}$ & \textcolor{blue}{2.799}$^{\pm{.072}}$  \\
    MLD ~\cite{mld} & $0.473^{\pm{.013}}$ & ${0.481}^{ \pm .003}$ & ${0.673}^{ \pm .003}$ & $0.772^{\pm{.002}}$ & $9.724^{\pm{.082}}$ & {2.413}$^{\pm{.079}}$  \\ 
 
    MotionDiffuse~\cite{motiondiffuse} & 0.630$^{\pm{.001}}$ & $0.491^{\pm{.001}}$ & $0.681^{\pm{.001}}$ & $0.782^{\pm{.001}}$ & $9.410^{\pm{.049}}$ & $1.553^{\pm{.042}}$   \\

    GMD \cite{gmd} & $0.212^{}$ & -&-&$0.670^{}$ & $9.440^{}$ & -  \\

    T2M-GPT~\cite{zhang2023t2mgpt}  & $0.116^{\pm{.004}}$ & $0.491^{\pm{.003}}$ & $0.680^{\pm{.003}}$ &$0.775^{\pm{.002}}$ & ${9.761}^{\pm{.081}}$ & 
    $1.856^{\pm{.011}}$  \\

    MotionGPT~\cite{jiang2024motiongpt} & $0.232^{\pm{.008}}$ & $0.492^{\pm{.003}}$ & $0.681^{\pm{.003}}$ &$0.778^{\pm{.002}}$ & \textcolor{red}{9.528}$^{\pm{.071}}$ & 
    $2.008^{\pm{.084}}$   \\

    ReMoDiffuse~\cite{zhang2023remodiffuse}  & \textcolor{blue}{0.103}$^{\pm{.004}}$ & \textcolor{blue}{0.510}$^{\pm{.005}}$ & \textcolor{blue}{0.698}$^{\pm{.006}}$ & \textcolor{blue}{0.795}$^{\pm{.004}}$ & $9.018^{\pm{.075}}$ & $1.795^{\pm{.043}}$  \\

    M2DM~\cite{m2dm}  & 0.352$^{ \pm .005}$ & 0.497$^{ \pm .003}$ & ${0.682}^{ \pm .002}$ & ${0.763}^{ \pm .003}$ & $9.926^{ \pm .073}$ & \textcolor{blue}{3.587}$^{ \pm .072} $\\

    Fg-T2M~\cite{wang2023fg}  & 0.243$^{ \pm .019}$ & 0.492$^{ \pm .002}$ & ${0.683}^{ \pm .003}$ & ${0.783}^{ \pm .002}$ & $9.278^{ \pm .072}$ & $1.614^{ \pm .049} $\\

    % AMD~\cite{priormdm}  & {0.204}$^{ \pm .031}$ & - & - & ${0.657}^{ \pm .006}$ & $9.476^{ \pm .077}$ & $1.356^{ \pm 

    \midrule
    
    StableMoFusion (Ours) & \textcolor{red}{0.098}$^{\pm{.003}}$ & \textcolor{red}{0.553}$ ^{\pm{.003}}$&  \textcolor{red}{0.748}$ ^{\pm{.002}}$ & \textcolor{red}{0.841}$ ^{\pm{.002}}$ & $9.748^{\pm{.092}}$ & $ 1.774^{\pm{.051}}$\\

    \bottomrule
  \end{tabular}
}
 
\end{table*}

\begin{table*}[t!]
 \caption{Quantitative results on the KIT-ML test set. The right arrow  $\rightarrow$ means the closer to real motion the better. \textcolor{red}{Red} and \textcolor{blue}{Blue} indicate the best and the second best result.}
\label{tab5: benchmark KIT}
  \centering
  
\resizebox{\linewidth}{!}{
  % \begin{tabular}{l c  m{1.8cm}<{\centering} c m{1.8cm}<{\centering}  m{1.5cm}<{\centering} cc}
  \begin{tabular}{lcccccc}
    \toprule
    \multirow{2}{*}{\centering Method} & 
    \multirow{2}{*}{\centering FID $\downarrow$} & 
    \multicolumn{3}{c}{R Precision$\uparrow$ }& 
    \multirow{2}{*}{\centering Diversity $\rightarrow$}  & 
    \multirow{2}{*}{\centering Multi-modality $\uparrow$}  
 \\
 \cline{3-5}
    && top1 & top2 & top3 & & \\
    \midrule
    Real Motion &  $0.031^{ \pm .004}$ &$0.424^{ \pm .005}$ & $0.649^{ \pm .006}$ & $0.779^{ \pm .006}$ &  $11.08^{ \pm .097}$  & - \\
    \midrule
    % JL2P ~\cite{ahuja2019language2pose} & $6.545^{ \pm .072}$  & $0.221^{ \pm .005}$ & $0.373^{ \pm .004}$ & $0.483^{ \pm .005}$  & $9.073^{ \pm .100}$ & - & - \\
    % Text2Gesture~\cite{bhattacharya2021text2gestures} & $12.12^{ \pm .183}$ & $0.156^{ \pm .004}$ & $0.255^{ \pm .004}$ & $0.338^{ \pm .005}$ & $9.334^{ \pm .079}$ & - & -\\
    % TEMOS ~\cite{petrovich2022temos} &  $3.717^{ \pm .051}$ & $3.417^{ \pm .019}$ & $10.84^{ \pm .100}$ & $0.532^{ \pm .034}$ &  $0.561^{+.007}$ & $0.687^{ \pm .005}$ & -  \\
     T2M ~\cite{t2m} & $2.770^{ \pm .109}$ & $0.370^{ \pm .005}$ & $0.569^{ \pm .007}$ & $0.693^{ \pm .007}$ & $10.91^{ \pm .119}$ & $1.482^{ \pm .065}$ \\
    MDM~\cite{MDM} & $0.497^{ \pm .021}$ &$0.164^{ \pm .004}$ & $0.291^{ \pm .004}$ & $0.396^{ \pm .004}$ &   $10.847^{ \pm .109}$ & {1.907}$^{ \pm .214}$  \\
    MLD~\cite{mld} & $0 . 4 0 4 ^ { \pm . 0 2 7 }$ & $0.390 ^{ \pm .008}$ & $0.609^{ \pm .008}$ & ${3.204}^{ \pm .027}$ & $10.80^{ \pm .117}$ & {2.192}$^{ \pm .071}$ \\
    
    MotionDiffuse~\cite{motiondiffuse} & $1.954^{ \pm .062}$ & $0.417^{ \pm .004}$ & $0.621^{ \pm .004}$ & $0.739^{ \pm .004}$  & $\textcolor{red}{11.10}^{ \pm .143}$ & $0.730^{ \pm .013}$ \\

    T2M-GPT~\cite{zhang2023t2mgpt}  & $0.514^{ \pm .029}$ & $0.416^{ \pm .006}$ & $0.627^{ \pm .006}$ & $0.745^{ \pm .006}$  & $10.921^{ \pm .108}$ & $1.570^{ \pm .039}$  \\

    MotionGPT~\cite{jiang2024motiongpt} & $0.510^{\pm{.016}}$ & $0.366^{\pm{.005}}$ & $0.558^{\pm{.004}}$ &$0.680^{\pm{.005}}$ & ${10.35}^{\pm{.084}}$ & \textcolor{blue}{2.328}
    $^{\pm{.117}}$  \\

    ReMoDiffuse~\cite{zhang2023remodiffuse}  & \textcolor{red}{0.155}$^{ \pm .006}$ & \textcolor{blue}{0.427}$^{ \pm .014}$ &\textcolor{blue}{0.641}$^{ \pm .004}$ & \textcolor{blue}{0.765}$^{ \pm .055}$ & $10.80^{ \pm .105}$ & $1.239^{ \pm .028}$  \\
    M2DM~\cite{m2dm}  & 0.515$^{ \pm .029}$ & 0.416$^{ \pm .004}$ & ${0.628}^{ \pm .004}$ & ${0.743}^{ \pm .004}$ & $11.417^{ \pm .97}$ & \textcolor{red}{3.325}$^{ \pm .37} $\\
    Fg-T2M~\cite{wang2023fg}  & 0.571$^{ \pm .047}$ & 0.418$^{ \pm .005}$ & ${0.626}^{ \pm .004}$ & ${0.745}^{ \pm .004}$ & $10.93^{ \pm .083}$ & $1.019^{ \pm .029} $\\
    % AMD~\cite{jing2024amd}  & ${0.233}^{ \pm .068}$ & -& - & ${0.401}^{ \pm .005}$ & $10.971^{ \pm .126}$ & 1.600$^{ \pm .174}$ \\

    \midrule
    
    StableMoFusion (Ours) &$\textcolor{blue}{0.258}^{\pm{.029}}$ & $\textcolor{red}{0.445} ^{\pm{.006}}$&  $\textcolor{red}{0.660} ^{\pm{.005}}$ & $\textcolor{red}{0.782} ^{\pm{.004}}$ & $\textcolor{blue}{10.936}^{\pm{.077}}$ & $ 1.362^{\pm{.062}}$  \\

    \bottomrule
  \end{tabular}
    }

\end{table*}

\section{Experiments}\label{sec:Comparison with the SOTA}

\subsection{Dataset and Evaluation Metrics} 
We use HumanML3D~\cite{t2m} and KIT-ML~\cite{plappert2016kit} datasets for our experiments. HumanML3D Dataset contains 14,646 motions and 44,970 motion annotations. KIT Motion Language Dataset contains 3,911 motions and 6,363 natural language annotations. 

The evaluation metrics can be summarized into four key aspects:
1) Motion Realism: Frechet Inception Distance (FID), which evaluates the similarity between generated and ground truth motion sequences using feature vectors extracted by a pre-trained motion encoder~\cite{t2m}. 
2) Text match: R Precision calculates the average top-k accuracy of matching generated motions with textual descriptions using a pre-trained contrastive model~\cite{t2m}.
3) Generation diversity: Diversity measures the average joint differences across generated sequences from all test texts. Multi-modality quantifies the diversity within motions generated for the same text.
4) Time costs: Average Inference Time per Sentence (AITS) ~\cite{mld} measures the inference efficiency of diffusion models in seconds, considering generation batch size as 1, without accounting for model or data loading time.

In all of our experiments, FID and R Precision are the principal metrics we used to conduct our analysis and draw conclusions.

\subsection{Implements Details}

For training, we use DDPM~\cite{ddpm} with $T=1,000$ denoising steps and variances $\beta_t$ linearly from 0.0001 to 0.02 in the forward process. And we use AdamW with an initial learning rate of 0.0002 and a 0.01 weight decay to train the sample-prediction model for 50,000 iterations at batch size 64 on an RTX A100. Meanwhile, the learning rate was reduced by 0.9 per 5,000 steps. On gradient descent, clip the gradient norm to 1. For CFG, setting c = $ \emptyset $ for 10\% of the samples.

For inference, we use the SDE variant of second-order DPM-Solver++~\cite{lu2022dpm++} with Karras Sigmas~\cite{ODE} in inference for sampling 10 steps. The scale for CFG is set to 2.5.

\subsection{Quantitative Results}

We compare our StableMoFusion with several state-of-the-art models
% , including T2M~\cite{t2m}, MDM~\cite{MDM}, MLD~\cite{mld}, MotionDiffuse ~\cite{motiondiffuse}, T2M-GPT~\cite{zhang2023t2mgpt}, MotionGPT~\cite{jiang2024motiongpt}, ReMoDiffuse~\cite{zhang2023remodiffuse}, M2DM~\cite{m2dm} and fg-T2M~\cite{wang2023fg}. 
% Quantitative comparisons of our method with these models
on the HumanML3D~\cite{t2m} and KIT-ML~\cite{plappert2016kit} datasets in ~\autoref{tab4: benchmark HumanML3D} and ~\autoref{tab5: benchmark KIT}, respectively. Most results are borrowed from their own paper and we run the evaluation 20 times and ± indicates the 95\% confidence interval. 

Our method achieves the state-of-the-art results in FID and R Precision (top k) on the HumanML3D dataset, and also achieves good results on the KIT-ML dataset: the best R Precision (top k) and the second best FID. This demonstrates the ability of StableMoFusion to generate high-quality motions that align
with the text prompts. On the other hand, while some methods excel in diversity and multi-modality, it's crucial to anchor these aspects with accuracy (R-precision) and precision (FID) to strengthen their persuasiveness. Otherwise, diversity or multimodality becomes meaningless if the generated motion is bad. Therefore, our StableMoFusion achieves advanced experimental results on two datasets and shows robustness in terms of model performance.

StableMoFusion$^*$ focuses on the real effect of footskate cleanup. Therefore, the timestep to begin cleaning footskate during inference depends on the motion and thus the StableMoFusion$^*$ doesn't apply to the evaluation process of ~\cite{t2m}.

\subsection{Qualitative Result}\label{sec:Qualitative result}
\begin{figure}[h!]
  \centering
   \includegraphics[width=\linewidth]{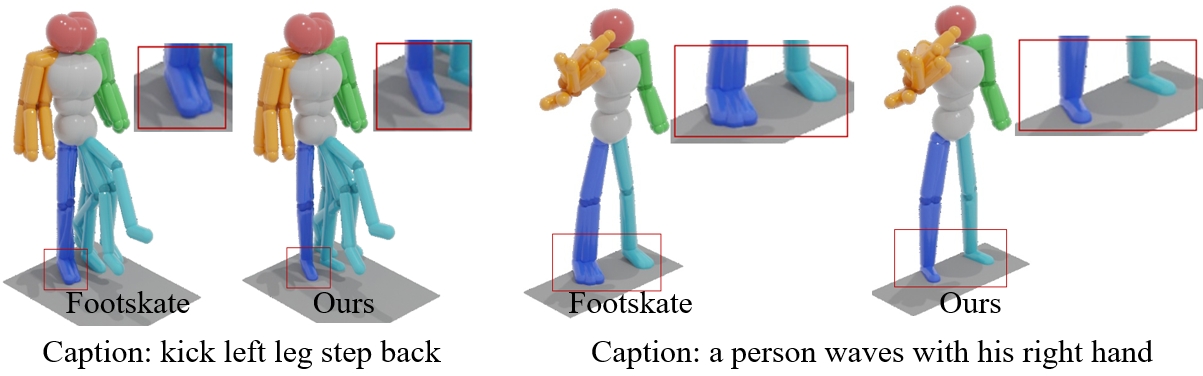}
   \caption{Visualization comparison results before and after our footskate cleanup. The red bounding box shows details of skating feet.
   }
   % , while our motions just show one.
   % }
   \label{fig:footskate_compare}
\end{figure}

\autoref{fig:footskate_compare} shows the qualitative results of our footskate cleanup method, StableMoFusion$^*$. The red bounding box of the footskate motion clearly has multiple foot outlines, whereas ours shows only one. The comparison graph shows the effectiveness of our method for cleaning footskate. Directly applying the footskate cleanup method of UnderPressure~\cite{Mourot22} to our motion would result in motion distortion, while our method effectively avoids such deformation. In our supplementary material, we will further present a comparison between our method and the UnderPressure method by videos to illustrate it.

\subsection{Inference Time}

We calculate AITS of StableMoFusion and ReMoDiffuse~\cite{zhang2023remodiffuse} with the test set of HumanML3D\cite{t2m} on Tesla V100 as MLD~\cite{mld} does, the other results of ~\autoref{fig:AITS} are borrowed from~\cite{mld}. For instance, MDM~\cite{MDM} with CFG requires 24.74s for average inference; MotionDiffuse~\cite{motiondiffuse} without CFG uses condition encoding cache and still requires 14.74s of average inference. While the MLD~\cite{mld} reduces the average inference time to 0.217s by applying DDIM50 in latent space, we find this approach lacks the ability to edit and control motion by manipulating the model input.

To tackle this, we employ 1) efficient sampler, 2) embedded-text cache, 3) parallel CFG computation, and 4) low-precision inference to reduce iteration counts and network latency. As shown in ~\autoref{fig:AITS}, our StableMoFusion significantly shortens the inference time and achieves higher performance within the original motion space.

However, it remains incontrovertible that StableMoFusion's inference speed trails behind that of MLD, and fails to meet the industry's real-time standard with an average inference time of 0.5s. Thus, our future work will focus on acceleration: the inference time of StableMoFusion is currently tied to the computation of the network, and we will further investigate how to scale down the model and how to reduce single-step latency in inference.

\subsection{Ablation}

\subsubsection{Network Architecture} \quad

We assess the validity of the improvements over the baseline for all network architectures proposed in \Cref{sec:Model Architecture} with uniform training and inference processes.

Specifically, we evaluate the baseline and adjusted model (+cross-attention) of Conv1D UNet, DiT, and RetNet. As shown in Table \ref{tab:ablation nertwork}, adding cross-attention notably boosts performance, highlighting its crucial role in enhancing the efficacy of conditional motion generation models.

Notably, \emph{Conv1D UNet + cross-attention} achieves superior results on HumanML3D, and with slight adjustments to group normalization (\emph{+ GroupNorm tweak}), also demonstrated robust performance on the KIT dataset. It proves that the GroupNorm tweak from sequence-wise normalization to feature-wise normalization on UNet is mainly useful for the datasets with dispersed length distributions, such as KIT-ML dataset.

\begin{table}[ht!]
 \caption{Quantitative comparison of architectural improvements in UNet, DiT and RetNet.}
 \label{tab:ablation nertwork}
 \centering
 \resizebox{\linewidth}{!}{
  \scriptsize
  \begin{tabular}{c|l| c| m{1.3cm}<{\centering}  }
    \toprule
    Dataset &Network  &   FID $\downarrow$& R Precision (top3) $\uparrow$ \\
    \toprule
    
    \multirow{7}{*}{\centering HumanML3D} &Conv1D UNet basline & 0.245 & 0.780 \\
    & \quad+ cross-attention & \textcolor{red}{0.074}  & 0.821  \\
    & \quad+ GroupNorm Tweak &  0.089 & \textcolor{red}{0.840}\\
    \cmidrule(lr){2-4}
    &DiT baseline &  0.884 & 0.711  \\
    
    & \quad+ cross-attention&  0.113 & 0.787\\
    
   \cmidrule(lr){2-4}
    
      & RetNet baseline & 1.673 & 0.740 \\
    & \quad+ cross-attention & 0.147 & 0.853  \\
    \midrule
     \multirow{2}{*}{\centering KIT-ML} & Conv1D UNet+ cross-attention & 0.658 & 0.756 \\
    & \quad+ GroupNorm Tweak & \textcolor{red}{0.237}  & \textcolor{red}{0.780}  \\
    \bottomrule
  \end{tabular}
}
\end{table}
% \subsubsection{Training Strategies}

% \begin{table}[]
%   \caption{The effect of EMA and CFG on CondUNet1D architectures.}
%   \label{tab3: progressive effect of inference}
%   \centering
%    \resizebox{0.8\linewidth}{!}{
%   \scriptsize
  
%   \begin{tabular}{l | c | m{1.7cm}<{\centering} }
%     \toprule
%     Method &   FID$\downarrow$& R Precision (top3)$\uparrow$ \\
%     \midrule
%      base(CondUNet1D) &  &\\
%     \quad+  EMA & 0.254 & 0.775 \\
%     \quad+  CFG& 0.089 & 0.839 \\
  
%     \bottomrule
%   \end{tabular}
%   }
% \end{table}

\subsubsection{Effective Inference} \quad

By using the SDE variant of second-order DPM-Solver++ with Karras sigma, the inference process of diffusion-based motion generation is able to significantly reduce the minimum number of iterations required for generation from 1000 to 10 while enhancing the quality of generated motions, approximately 99\% faster than the original inference process, as shown in ~\autoref{tab3: progressive effect of inference}.

The application of embedded text caching and parallel CFG further reduces the average inference time by about 0.3 seconds and 0.15 seconds, respectively. Our experiments also show that reducing the computational accuracy of the motion-denoising model by half, from FP32 to FP16, does not adversely affect the generation quality. This suggests that 32-bit precision is redundant for the motion generation task.

\begin{table}[h!]
  \caption{The progressive effect of each efficient and training-free trick of StableMoFusion in inference process.   }
  \label{tab3: progressive effect of inference}
  \centering
   \resizebox{\linewidth}{!}{
  \scriptsize
  
  \begin{tabular}{l | c | m{1.3cm}<{\centering}| c |  m{1.4cm}<{\centering} }
    \toprule
    Method &   FID$\downarrow$& R Precision (top3)$\uparrow$ & AITS$\downarrow$& Inference Steps$\downarrow$ \\
    \midrule
    base (DDPM1000)&  1.251 & 0.760 & 99.060 & 1000 \\
    \quad+  Efficient Sampler &  0.076  & 0.836 & 1.004(\textcolor{blue}{-99\%}) & 10\\
    \quad+  Embedded-text Cache &  0.076  & 0.836 & 0.690(\textcolor{blue}{-31\%}) & 10\\
    \quad+  Parallel CFG &  0.076 & 0.836 & 0.544\textcolor{blue}{(-21\%)} & 10 \\
    \quad+  FP16 &  0.076 & 0.837& 0.499\textcolor{blue}{(-8\%)} & 10 \\
    
    \bottomrule
  \end{tabular}
  }
\end{table}

\section{Conclusion}
In this paper, we propose a robust and efficient diffusion-based motion generation framework, StableMoFusion, which uses Conv1DUNet as a motion-denoising network and employs two effective training strategies to enhance the network's effectiveness, as well as four training-free tricks to achieve efficient inference. 
Extensive experimental results show that our StableMoFusion performs favorably against current state-of-the-art methods. Furthermore, we propose effective solutions for time-consuming inference and footskate problems, facilitating diffusion-based motion generation methods for practical applications in industry.



 \bibliographystyle{plain} 
 \balance
 \bibliography{ref}

% \newpage
% \appendix
% \section*{Appendix}
% \section{Appendix A: Additional Information}

\end{document}